\documentclass[12pt, onecolumn]{article}
\usepackage{booktabs}  
\usepackage{diagbox}   
\usepackage{multirow}  
\usepackage{fullpage}
\usepackage{amssymb}
\usepackage{graphicx}
\usepackage{authblk}
\usepackage{atbegshi}
\usepackage{placeins}
\usepackage{xcolor}
\usepackage{soul}
\usepackage[colorlinks,linkcolor=red,anchorcolor=green,citecolor=green]{hyperref}
\usepackage[absolute]{textpos}
\title{LM-Net: A Light-weight and Multi-scale  Network for Medical Image Segmentation}

\author[1]{Zhenkun Lu}
\author[1]{Chaoyin She\thanks{ Equal contribution} }
\author[2]{Wei Wang}
\author[3]{Qinghua Huang\thanks{Corresponding author	 
	
\texttt{\url{ https://github.com/Asunatan/LM-Net}}
}}

\affil[1]{College of Electronic Information, Guangxi Minzu University, Nanning, China}
\affil[2]{Department of Medical Ultrasonics, Institute of Diagnostic and Interventional Ultrasound, The First Affiliated Hospital of Sun Yat-Sen University, Guangzhou, China}
\affil[3]{School of Artificial Intelligence, Optics and Electronics (iOPEN), Northwestern Polytechnical University, Xi’an, China.}

\date{} 
\begin{document}
\maketitle
\begin{abstract}
	Current medical image segmentation approaches have limitations in deeply exploring multi-scale information and effectively combining local detail textures with global contextual semantic information. This results in over-segmentation, under-segmentation, and blurred segmentation boundaries. To tackle these challenges, we explore multi-scale feature representations from different perspectives, proposing a novel, lightweight, and multi-scale architecture (LM-Net) that integrates advantages of both Convolutional Neural Networks (CNNs) and Vision Transformers (ViTs) to enhance segmentation accuracy. LM-Net employs a lightweight multi-branch module to capture multi-scale features at the same level. Furthermore, we introduce two modules to concurrently capture local detail textures and global semantics with multi-scale features at different levels: the Local Feature Transformer (LFT) and Global Feature Transformer (GFT). The LFT integrates local window self-attention to capture local detail textures, while the GFT leverages global self-attention to capture global contextual semantics. By combining these modules, our model achieves complementarity between local and global representations, alleviating the problem of blurred segmentation boundaries in medical image segmentation. 
	To evaluate the feasibility of LM-Net, extensive experiments have been conducted on three publicly available datasets with different modalities. Our proposed model achieves state-of-the-art results, surpassing previous  methods, while only requiring 4.66G FLOPs and 5.4M parameters. These state-of-the-art results on three datasets with different modalities demonstrate the effectiveness and adaptability of our proposed LM-Net for various medical image segmentation tasks.
	
Keywords: Medical image segmentation, Multi-scale feature representation, Local Feature Transformer, Global Feature Transformer

\end{abstract}

\section{Introduction}

Medical images play a vital role in diagnosing and evaluating diseases by providing insights into physiological changes in the human body. In recent years, artificial intelligence (AI) has made remarkable progress in medical diagnosis, especially in medical image segmentation. As an indispensable process, medical image segmentation accurately delineates lesion contours and identifies lesion locations, thus offering critical diagnostic information for further pathological analysis. Therefore, it is an essential component of computer-aided diagnosis (CAD) systems used by clinicians.

Convolutional neural networks (CNNs) have gained considerable attention for medical image segmentation, owing to their capacity for learning complex features and effective scalability. Among the most successful CNN-based models is Unet \cite{U-net}, which employs a symmetric encoder-decoder structure to capture semantic features and demonstrate exceptional performance on various medical image segmentation tasks \cite{huang2023review}.
The success of Unet has spurred the development of numerous variants \cite{resunet++,unet++,Unet3+,ResUnet} and extensions to 3D medical image analysis \cite{3D-Unet,V-Net,Fuse-UNet}. However, these models are limited by their small receptive fields, which restrict their capacity to model long-range contextual semantic information. To address this limitation, researchers have incorporated attention mechanisms  \cite{SE,Pyramidfeatureattentionnetwork,cbam,Attentionu-net,luo2022segmentation,huang2022anatomical,huang2022evaluation,huang2023nag,zhou2023bsmnet} into the architecture, resulting in improved model robustness.
 
 The recent success of Vision Transformer (ViT) \cite{ViT} in image-level prediction tasks has inspired researchers to explore its combination with medical image segmentation \cite{MedT,Swin-unet,MCTrans}. ViT, a Transformer-based architecture, has achieved state-of-the-art performance in image classification tasks by modeling global spatial relationships and long-range dependencies. However, its high computational cost for large images and lack of local inductive bias can limit effectiveness for high-resolution visual tasks. A typical solution is to combine CNNs and Transformers, leveraging the strengths of both approaches. These hybrid models \cite{transunet,uctransnet} have shown promising results on various medical image segmentation tasks.
 
 Despite these advancements, the aforementioned methods suffer from various limitations: (1) They have not effectively explored multi-scale information that is crucial in achieving accurate predictions on dense image segmentation tasks. (2) There is inadequate integration of local detail textures and global contextual semantics, resulting in issues like over/under-segmentation and blurred boundaries.
 (3) Most models are typically not lightweight, posing significant challenges for integration into CAD systems that use embedded devices with limited resources.
  Consequently, further research efforts need to be directed towards addressing these issues to enhance the feasibility of applying deep learning techniques for medical image segmentation in real-world scenarios. 
 
  To overcome these limitations, we propose a novel, lightweight, and multi-scale architecture called LM-Net. LM-Net comprises three meticulously designed modules: the multi-branch module, Local Feature Transformer (LFT), and Global Feature Transformer (GFT). LM-Net explores multi-scale features from two primary perspectives: (1) The multi-branch module has multiple parallel depth-wise convolutions to effectively capture multi-scale features at a same level. More importantly, structural re-parameterization \cite{Repvgg} can be used to decouple the training phase and the inference phase, ensuring that the multi-branch module has consistent number of parameters and computational cost compared to the original block during the inference phase. (2) LFT employs local window self-attention to extract fine-grained textures while GFT leverages global self-attention to capture global contextual semantics.
The global features characterize the overall lesion region to distinguish it from surrounding tissue. Meanwhile, local features focus on refining boundaries and delineating the lesion precisely.
Both modules are designed to take the encoder's feature pyramid as input, which allows LM-Net can effectively utilize rich contextual multi-scale features at different levels. 

Our proposed model achieves state-of-the-art performance on the Kvasir-SEG dataset \cite{Kvasir-seg} with 94.09\% mDice and 89.12\% mIoU, outperforming prior best methods like ESFPNet-L (92.5\% mDice and 87.5\% mIoU) \cite{ESFPNet}, SSFormer-L (93.57\% mDice, 89.05\% mIoU) \cite{SSFormer} and  FCBFormer (93.85\% mDice, 89.03\% mIoU) \cite{FCBFormer}. Superior results are also obtained on the LGG Segmentation dataset \cite{LGG1,LGG2} and breast ultrasound Images dataset \cite{BUIS}.

 In short, our contributions are summarized as follows:
\begin{itemize}

\item We propose a lightweight multi-branch module that can readily use structural re-parameterization to transform the model structure equivalently, without incurring additional inference-time costs.

\item  We introduce two plug-and-play modules, GFT and LFT, to integrate local textures and global semantics for enhanced representation learning, alleviating issues like blurred boundaries.

\item Multi-scale features are explored from two perspectives, the multi-branch module extracts multi-scale features at a same level, whereas LFT and GFT capture multi-scale information at different levels to minimize the semantic gap between high-level and low-level information.
\item The proposed model demonstrates promising performance on various medical image data with different modalities, which suggests its potential as a reliable and versatile medical image segmentation model.

\end{itemize}

\section{Related Work}
\subsection{Convolution neural networks}
Current semantic segmentation models based on convolutional neural networks (e.g., FCN \cite{FCN}) generally employ an encoder-decoder architecture: (1) the encoder extracts high-level semantics and low-level texture features by progressively downsampling the input image. (2) the decoder aggregates these features and converts them into a final dense prediction through layer-wise upsampling. 

To explore potential semantic information, this architecture typically requires a robust encoder (e.g., VGG \cite{VGG}). Tan et al. \cite{Efficientnet} systematically analyzed the influence of network width and depth, proposing a simple yet efficient composite scale method that extends pre-existing baseline convolutional neural networks while maintaining model validity. He et al. \cite{RESNET} introduced shortcut connections to address the challenges associated with training deeper neural networks over standard architectures. Chen et al. \cite{Deeplab,DeeplabV3+,AtrousConvolution} adopted the atrous convolution to expand receptive fields and proposed ASPP which can effectively capture target and context information at multiple scales. Another approaches leverage attention mechanisms integrated with convolutional neural networks. Squeeze-and-Excitation (SE) \cite{SE} improves the quality of the representation by modeling the interdependence between feature mapping channels. Pyramid Feature Attention Network \cite{Pyramidfeatureattentionnetwork} has a contextual-aware ability to capture rich semantic features on high-level feature maps. CBAM \cite{cbam} sequentially applies channel and spatial attention modules to complement each other and enhance representation capabilities. Dual Attention Network(DANet) \cite{DANet} adaptively aggregates long-range contextual semantic information via parallel spatial and channel attention modules, improving the discrimination of feature representation. 

Convolutional neural networks (CNNs) have been increasingly utilized in the medical field \cite{huang2023extraction,huang2022anatomical,huang2021dense}, particularly in  computer-aided diagnosis systems. U-shaped segmentation models, including Unet and its variants like ResUnet \cite{ResUnet}, ResUnet++\cite{resunet++}, Unet++\cite{unet++}, Unet3+\cite{Unet3+}, and AttentionUnet \cite{Attentionu-net}, have shown promising results in medical image analysis tasks. These models can effectively segment and identify regions of interest in medical images to assist diagnosis and treatment decisions. The U-shaped structure has also been used in 3D medical image analysis with models such as 3D-Unet\cite{3D-Unet} and V-Net \cite{V-Net}, as well as other domains like road scene segmentation (e.g., Segnet\cite{Segnet}) and remote sensing (e.g., ResUnet-a \cite{ResUNet-a}).

Current medical image segmentation methods have primarily focused on network depth and width optimization or employing attention mechanisms to capture interchannel feature dependencies. However, these approaches do not fully exploit the potential of multi-scale information. Therefore, we propose a novel approach that explores multi-scale features through two primary perspectives: (1) incorporating lightweight multi-branch modules that simultaneously capture multi-scale features at the same level. (2) introducing LFT and GFT to extract multi-scale information at different levels. 
\subsection{Vision Transformers}
Recent advancements in natural language processing (NLP) have inspired researchers to apply Transformer-based methods to computer vision tasks \cite{huang2023novel}. The Vision Transformer (ViT) is a noteworthy example of this, as it directly applies a standard Transformer to process images. Specifically, ViT divides an input image into a series of patches and relies on Multi-Head Self-Attention (MHSA) \cite{MHSA} to capture patch-to-patch dependencies, demonstrating impressive speed-accuracy trade-off in image classification tasks.

 However, the quadratic computational cost of self-attention on larger image sizes remains a challenge for downstream tasks such as object detection and semantic segmentation. To overcome this limitation, recent works have proposed improvements to ViT's applicability for dense prediction tasks. For instance, Zheng et al.\cite{SERT} presents a novel approach to semantic segmentation through sequence-to-sequence prediction, while Ranftl et al. \cite{DPT} employs a dense neural network architecture that effectively leverages ViT to produce fine-grained and globally consistent features for dense prediction tasks.
     
  Vision Transformer has also found application in the medical field, particularly in image segmentation. TransUnet \cite{transunet} is the first composite U-shaped structure that combines the benefits of CNN and ViT. Swin-Unet \cite{Swin-unet} utilizes a shift window scheme to construct a U-shaped hierarchical architecture, which is the first pure Transformer-based architecture in the medical domain. UCTransNet \cite{uctransnet} adopts CCT and CCA modules as a skip-connection between the encoder and decoder to address the semantic gap problem. 
However, ViT suffers from a lack of local inductive bias, making local detail texture extraction challenging. Additionally, ViT models typically require extensive pre-training on very large datasets like ImageNet \cite{imagenet}. To mitigate these limitations, we propose the GFT and LFT modules to explore global contexts and local details respectively, without relying on heavy pre-training.

\section{Method}
\subsection{Overall Architecture }
\begin{figure}[!ht]
	\centering
	\includegraphics[width=1\linewidth]{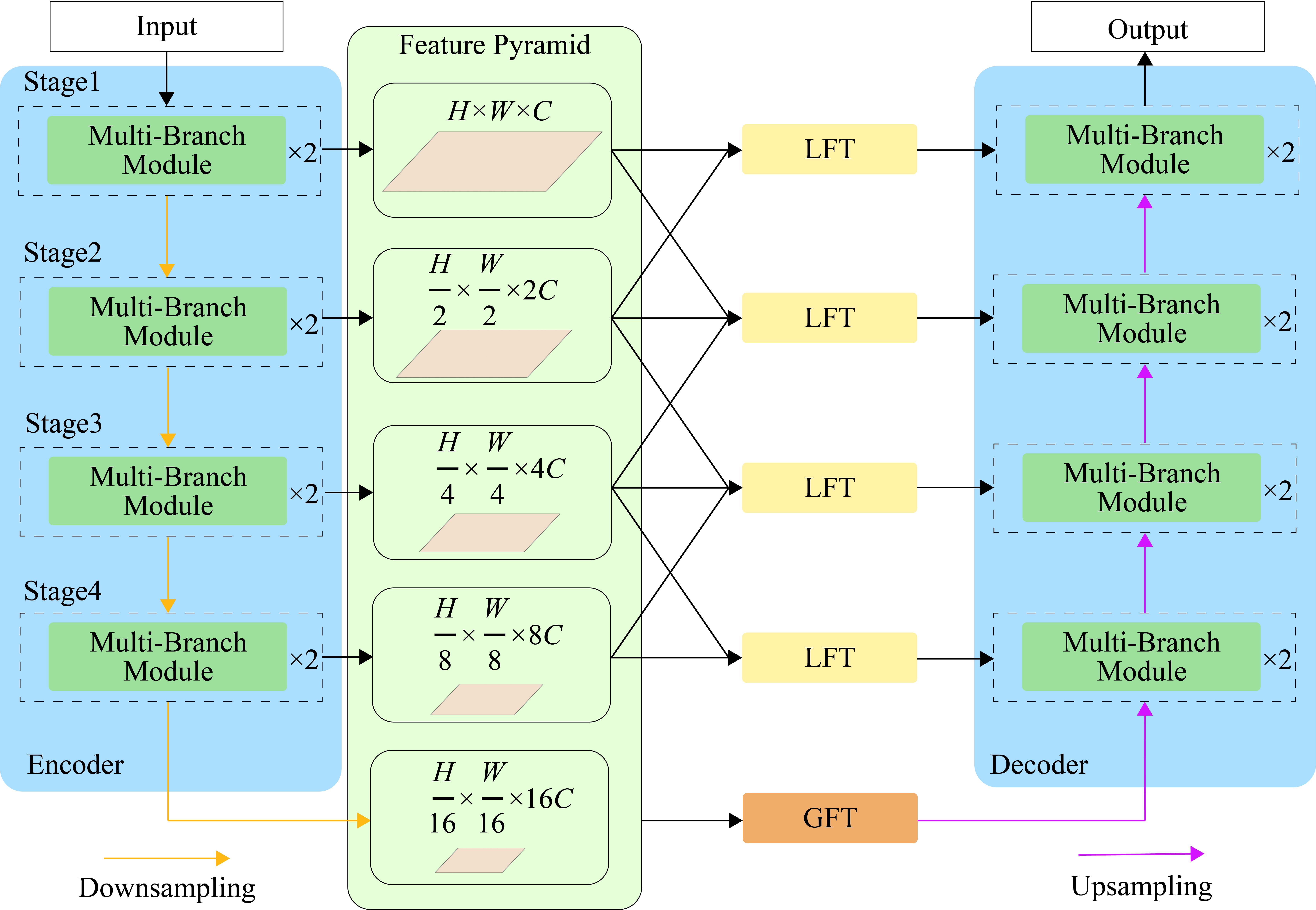}
	\caption{The overall architecture of the LM-Net. LM-Net is a symmetric hierarchical structure, which consists of an encoder, decoder, and skip-connection. The encoder and decoder are composed of multi-branch modules. LFT captures local detail textures, while GFT explores global semantics. The decoder integrates local detail textures and global semantics.}
	\label{Figure 1}
\end{figure}

Medical imaging segmentation targets often exhibit inconsistencies leading to under/over-segmentation, and blurred boundaries, posing significant challenges for accurate diagnosis and treatment planning. To mitigate these issues, we propose LM-Net, a novel, lightweight, and multi-scale architecture. As illustrated in Figure \ref{Figure 1}, LM-Net adopts a U-shaped design with an encoder-decoder structure. It explores multi-scale features from varying perspectives, integrating both local and global information to achieve better segmentation quality than methods relying on either local or global cues alone.

Existing segmentation approaches incur substantial computational costs and parameters when expanding receptive fields or employing self-attention, making them less suitable for clinical computer-aided diagnosis (CAD) systems. Moreover, they often neglect multi-scale representations critical for dense predictions. Therefore, developing a lightweight model that can effectively extract multi-scale features is crucial.
To address this need, we introduce a lightweight multi-branch module which serves as the fundamental module for both the encoder and decoder. Using parallel convolutions with distinct kernel sizes, the multi-branch module can capture potential multi-scale representation information at the same level. As depicted in Figure \ref{Figure 1}, the LM-Net encoder comprises four stages, each containing two multi-branch modules. Downsampling operations between consecutive stages halve the spatial dimensions while doubling the number of channels. Specifically, for an image with dimension of $ H\times W\times 3 $, stage $i$ outputs feature maps of dimension $\frac{H}{2^{i-1}} \times \frac{W}{2^{i-1}} \times 2^{i-1}C$.
To address the issues of under/over-segmentation and blurred boundaries, we propose two novel modules: Local Feature Transformer (LFT) and Global Feature Transformer (GFT), which integrate local and global self-attention into the architecture. These two modules are used to extract local detail texture and global semantics, respectively.  Specifically, we construct a series of local feature pyramids using feature maps from adjacent stages in the encoder, and then apply the LFT module to capture local details while reducing semantic gaps. Furthermore, we create a global feature pyramid using feature maps from all stages in the encoder as input to the GFT module. This enables the model to exploit multi-scale features at different levels to learn long-range dependencies and accurately localize lesions.

 Finally, the decoder fuses local detail texture and global semantic context to generate the final segmentation result. This is a layer-by-layer late fusion strategy that allows efficient aggregation of information from different scales and abstraction levels.

\subsection{Multi-branch Module }
\begin{figure}[h]
	\centering
	\includegraphics[width=1\linewidth]{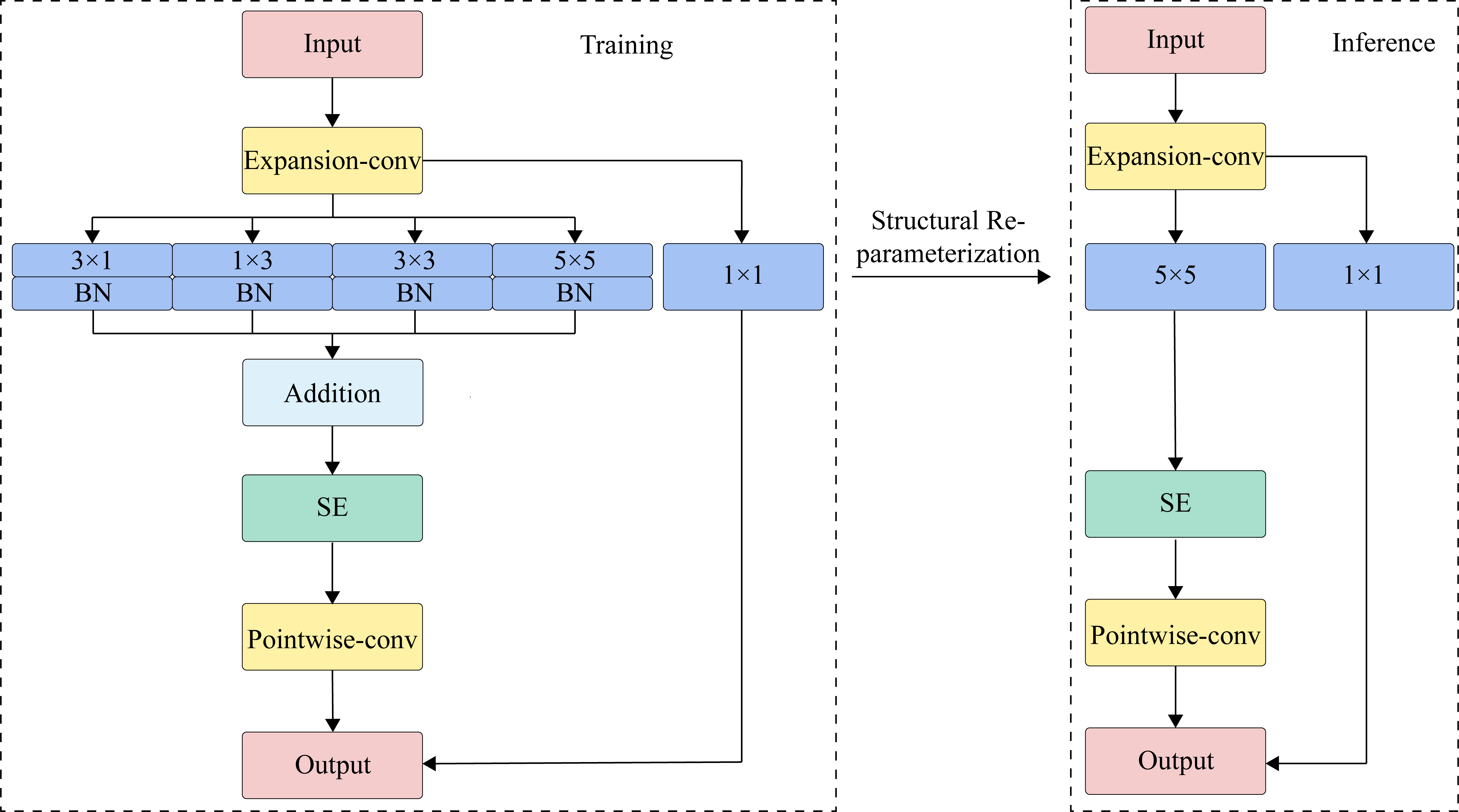}
	\caption{Overview of multi-branch module. In the training phase, it contains four branches. In the inference phase, we equivalently convert the parameters of convolution and BN  so that it maintains the same structure as a original block.}
	\label{Figure 2}
\end{figure}

Medical image segmentation often necessitates the utilization  of multi-scale features to handle targets of varying sizes. However, current segmentation methods do not effectively leverage these features, making it difficult to accurately segment targets with different scales. To address this limitation, we propose a lightweight multi-branch module that can efficiently extract multi-scale features at the same level. As illustrated on the left in Figure \ref{Figure 2}, the multi-branch module comprises an $ 1\times1 $ expansion convolution, followed by depth-wise separable convolutions and a Squeeze-and-Excitation (SE) \cite{SE} layer. The depth-wise separable convolution consists of two independent layers: a depth-wise convolution to extract spatial features and a  $ 1\times1 $ point-wise convolution for channel-wise feature learning. The SE layer is added between the depth-wise and point-wise convolution to enhance the quality of feature representations by learning the interdependence between feature mapping channels. The key modification is in the depth-wise convolution branch, which contains four parallel layers, each with a convolution and a batch normalization (BN) \cite{BN}. The kernel size of these convolutions is $ 3\times1 $, $ 1\times3 $, $ 3\times3 $, and $ 5\times5 $, respectively, enabling the module to capture multi-scale contextual information at the same level through parallel convolutions. This is attributed to the distinct receptive fields of different-sized convolution kernels. Additionally, a convolution layer is added to serve as a shortcut-connection, ensuring the consistency of feature channels and facilitating gradient backpropagation optimization, thereby simplifying network training. In summary, the proposed lightweight multi-branch module allows efficient extraction of multi-scale features at the same level.
This multi-branch module can easily decouple the training and inference stages through structural re-parameterization. We now provide a detailed explanation of this technique.
 Let $ F\in \mathbb{R} ^{C\times H \times W} $ ( excluding the batch size ) denote a feature map with $ C $ channels and spatial resolution $ H\times W $. The convolutional kernel is represented  as  $ K\in \mathbb{R} ^{C\times h \times w\times D}  $, where $ C $, $ D $ and $ h \times w $ represent the number of input channels, output channels and  kernel size, respectively. The corresponding output of the convolution operation between $ F $ and $ K $ is expressed as  $ O\in \mathbb{R} ^{D\times H^{'}  \times W^{'} } $, where $ H^{'} $ and $ W^{'} $ depend on $ K $. This can be formulated as:
\begin{equation}\label{equation1}
	O= F\otimes K
\end{equation}
where $ \otimes $ denotes the convolution operator. Batch normalization is commonly used with CNNs to reduce overfitting and accelerate training \cite{BN}. With batch normalization,  the output in Eq. \ref{equation1} becomes:
\begin{equation}\label{equation2}
	O= (F  \otimes  K - \mu )\frac{\eta }{\delta  } +\beta 
\end{equation}
where $ \eta $ and $ \beta $ represent the learnable scaling factor and bias term, while $ \mu $ and $ \delta $ denote the average value and standard deviation of batch normalization, respectively. The fundamental principle of structural re-parameterization is to leverage the homogeneity and additivity of convolution operations:

\begin{equation}\label{equation3}
	 F  \otimes ( \lambda K )=\lambda(F\otimes K),\forall \lambda\in \mathbb{R} 
\end{equation}
\begin{equation}\label{equation4}
 F  \otimes K_{1} +  F  \otimes K_{2} = F\otimes (K_{1}+K_{2})
\end{equation}
The homogeneity allows fusing BN into the convolution during inference by converting $ K $ to $ K^{'} $ with extra bias $ b $, yielding the same output as Eq. \ref{equation2}:

\begin{equation}\label{equation5}
O = F\otimes  K^{'} +b 
\end{equation}
\begin{equation}\label{equation6}
K^{'}=\frac{\eta }{\delta  }K,b=-\mu \frac{\eta }{\delta  }+\beta 
\end{equation}
As depicted in Figure \ref{Figure 2}, the structural re-parameterization technique allows us to easily decouple the training and the inference stage of the multi-branch module. During inference, the four branches merge into one single standard convolution by element-wise adding the kernel parameters and the four bias vectors at the corresponding positions. Consequently, this process yields Eq. \ref{equation7} and \ref{equation8} as follows:
\begin{equation}\label{equation7}
K^{'}=\frac{\eta_{1}  }{\delta_{1}  }K_{1} +\frac{\eta_{2}  }{\delta_{2}  }K_{2}+\frac{\eta_{3}  }{\delta_{3}  }K_{3}+\frac{\eta_{4}  }{\delta_{4}  }K_{4}
\end{equation}
\begin{equation}\label{equation8}
b=-\mu_{1}\frac{\eta_{1}}{\delta_{1}}-\mu_{2}\frac{\eta_{2}}{\delta_{2}}-\mu_{3}\frac{\eta_{3}}{\delta_{3}}-\mu_{4}\frac{\eta_{4}}{\delta_{4}}+\beta_{1} +\beta_{2} +\beta_{3} +\beta_{4} 
\end{equation}
This multi-branch design can be easily extended. For $n$ branches, the general formulation is:
\begin{equation}\label{equation9}
K=\Sigma_{i=1}^{n}\frac{\eta_{i}}{\delta_{i}}K_{i} ,b=\Sigma_{i=1}^{n}-\mu_{i}\frac{\eta_{i}}{\delta_{i}}+\beta_{i} 
\end{equation}
The multi-branch module incorporates four parallel branches, allowing it to capture multi-scale information while maintaining computational complexity equivalent to a single convolutional layer. This is achieved by having each branch use a different kernel size ($3\times1$, $1\times3$, $3\times3$, $5\times5$) to extract features at different scales simultaneously. As a result, the module enables more efficient inference since the combined computational cost of the four branches is comparable to a single convolution. During training, the branches are optimized independently with their own sets of parameters. Separately updating the branches not only improves efficiency but also facilitates training deeper, more complex networks. Overall, the proposed multi-branch design enhances model efficiency for both inference and training in multi-scale feature learning.
\subsection{Global Feature Transformer and Local Feature Transformer}
\textbf{GFT}: Some pioneering works \cite{UperNet,FPN,PANet,PSP,SPP} have demonstrated the effectiveness of feature pyramids for dense prediction tasks. Inspired by UperNet \cite{UperNet}, we develop a Transformer-based module called GFT to fuse feature pyramids from the encoder as a bridge between the encoder and decoder. By leveraging multi-scale features in the feature pyramid hierarchy, the GFT module can effectively capture global contextual semantics at different scales. To optimize the feature pyramid, we designed a convolution embedding method to replace standard patch embedding. The convolutional embedding includes reshaping, convolution, concatenation, and flattening operations.
\begin{figure}[h]
	\centering
	\includegraphics[width=1\linewidth]{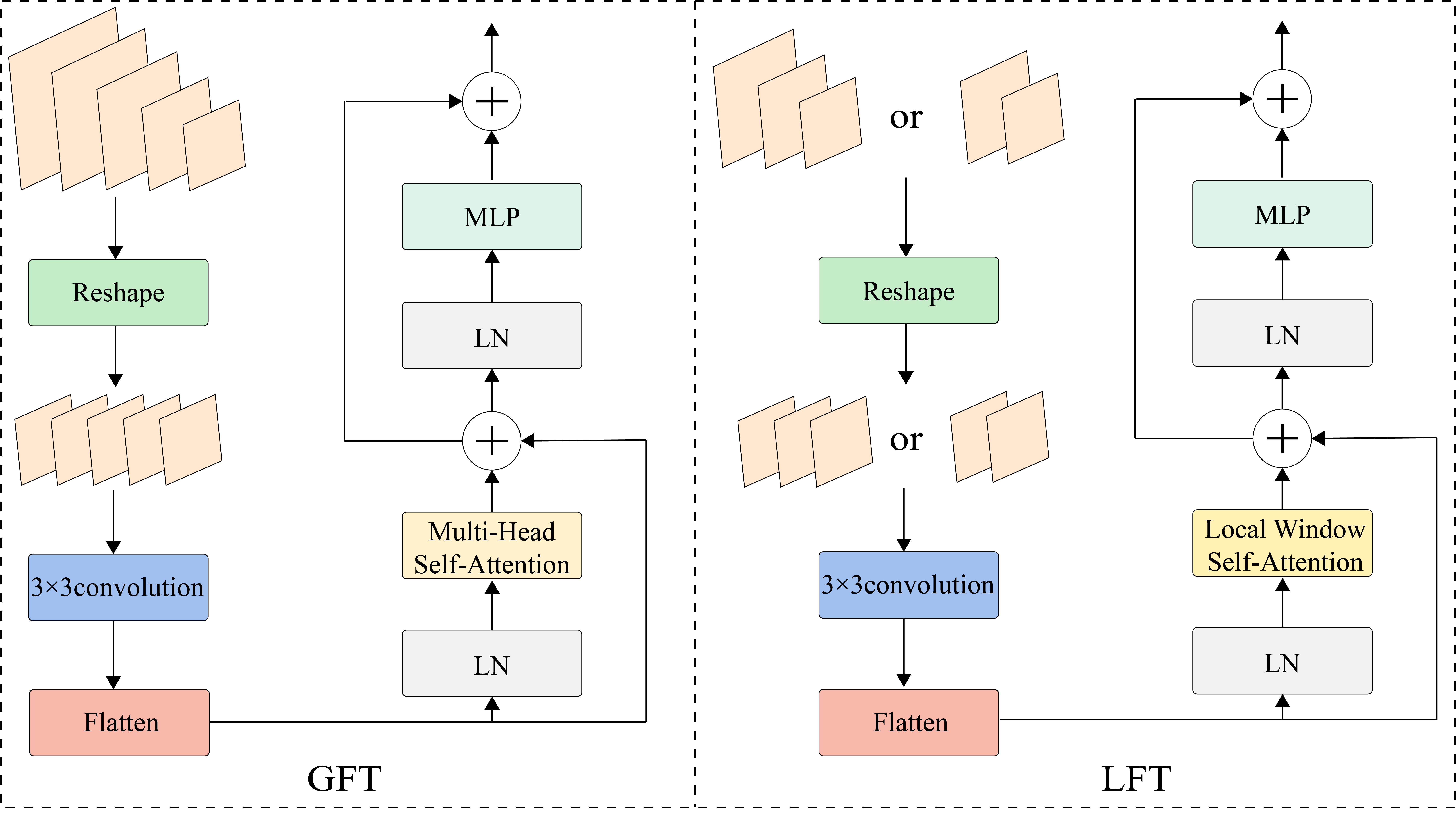}
	\caption{GFT module and LFT module. On the left is GFT and on the right is LFT, both of which use a feature pyramid as input. GFT uses global self-attention to capture global semantics, while LFT uses local window self-attention to capture local details.}
	\label{Figure 3}
\end{figure}
As illustrated on the left side of Figure \ref{Figure 3}, the GFT module takes a global feature pyramid with five feature maps from different stages of the encoder as input. These input feature maps of different scales are reshaped to a target size and aggregated via a $ 3\times 3 $ convolution. The output of the convolution is directly flattened  into a vector, which is then fed into the Transformer. This approach reduces the computational load and addresses the lack of local inductive bias in Transformers, improving performance in capturing fine-grained details. 

To construct the feature pyramid, we analyze encoder feature maps $ x_{1},x_{2},x_{3},x_{4} $ and $ x_{5} $ with resolution sizes of $ 1,  1/2, 1/4, 1/8, $ and $ 1/16 $, relative to the input image. The feature pyramid $ X=\left \{x_{1},x_{2},x_{3},x_{4},x_{5}\right \} $  is then downsampled to the target size $ \frac{H}{S} \times \frac{W}{S} $ (where $ S $ is the stride relative to the input image, e.g.,
 $ \frac{H}{16} \times \frac{W}{16} $). The reshaped pyramid is concatenated channel-wise and convolved with a $3 \times 3$ kernel to aggregate multi-scale features. Finally, we flatten the new feature map to a vector (sequence) $ X^{'}  $:
\begin{equation}\label{equation10}
X^{'} =Flatten(Conv(Concat(Reshape(X))))
\end{equation}

To produce semantic information with global contextual awareness, the flattened feature map $ X^{'}  $ is fed into a Transformer module. This module consists of a multi-head self-attention (MHSA), a 2-layer MLP, two LayerNorm (LN) layers, and two residual connections. The processing of the Transformer module can be expressed as:
\begin{equation}\label{equation11}
	X_{a} =X^{'} +MHSA(LN(X^{'}))
\end{equation}
\begin{equation}\label{equation12}
	X_{o} =X_{a}  +MLP(LN(X_{a}))
\end{equation}
Particularly, the multi-head self-attention(MHSA) includes three components: query($ Q $), key($ K $), and value($ V $). Each of the $ Q $,  $ K $, and $ V $ head has equal dimension, which is determined by the length of the input sequence and channels. The self-attention mechanism is estimated as :
\begin{equation}\label{equation13}
Att=Softmax(\frac{QK^{T}}{\sqrt{d} } )V
\end{equation}
where $ d=C/M $ is the head dimension and $ M $ is the number of heads. The output  $ X_{o} \in \mathbb{R}^{\frac{HW}{S^{2}}\times C} $ is reshaped to $ C \times \frac{H}{S} \times \frac{W}{S}  $, and then transmitted to the decoder.
\\
\textbf{LFT}: Similar to GFT, we have also designed a lightweight and efficient module, LFT. As shown on the right side of Figure \ref{Figure 3}, LFT is built upon a simple local self-attention mechanism \cite{NAT} that demonstrates a local inductive bias and linear complexity. To minimize the semantic gap between high-level and low-level information, the output of adjacent stages within the encoder are combined to build a local feature pyramid as the input of the LFT. The number of layers in the local feature pyramid depends on the stage at which the LFT is located: the first and fourth stages consist of two layers, while the second and third stages contain of three layers. The local feature pyramid is reshaped to the target size according to the stage of the LFT, and then concatenated along the channel dimension, aggregated via a $3\times 3$ convolution. Finally, the aggregated features are fed into a Transformer for further processing.

 This approach has two advantages: (1) The local feature pyramid aggregates rich spatial and semantic information from different scales, which is critical for medical image segmentation. (2) The LFT’s  local inductive bias enables it to provide sufficiently fine edge and texture information which can help the decoder recover the full spatial resolution. The final prediction result is obtained by the decoder aggregating the local texture information from the LFT and the global semantic information from the GFT. As shown in Figure \ref{Figure 1}, this process employs a hierarchical post-fusion strategy. By leveraging the advantages of GFT and LFT, LM-Net demonstrates the effectiveness of the proposed method in medical image segmentation and achieves state-of-the-art performance on several benchmark datasets.

\subsection{Multi-scale features}
In this section, we investigate multi-scale features from two distinct perspectives. The first perspective involves using convolution kernels of varying sizes on the same level feature map to capture multi-scale information, commonly known as multi-scale convolution. This technique extracts feature information across multiple scales by employing convolution kernels of different dimensions, thus better capturing the target object's varying scales. The second perspective, similar to UperNet in existing research, aims to explore multi-scale information by utilizing feature maps at different levels. This method constructs a feature pyramid using feature maps from different encoder layers or stages, and then fuses these to achieve a comprehensive multi-scale feature representation.
 \begin{figure}[!ht]
	\centering
	\includegraphics[width=1\linewidth]{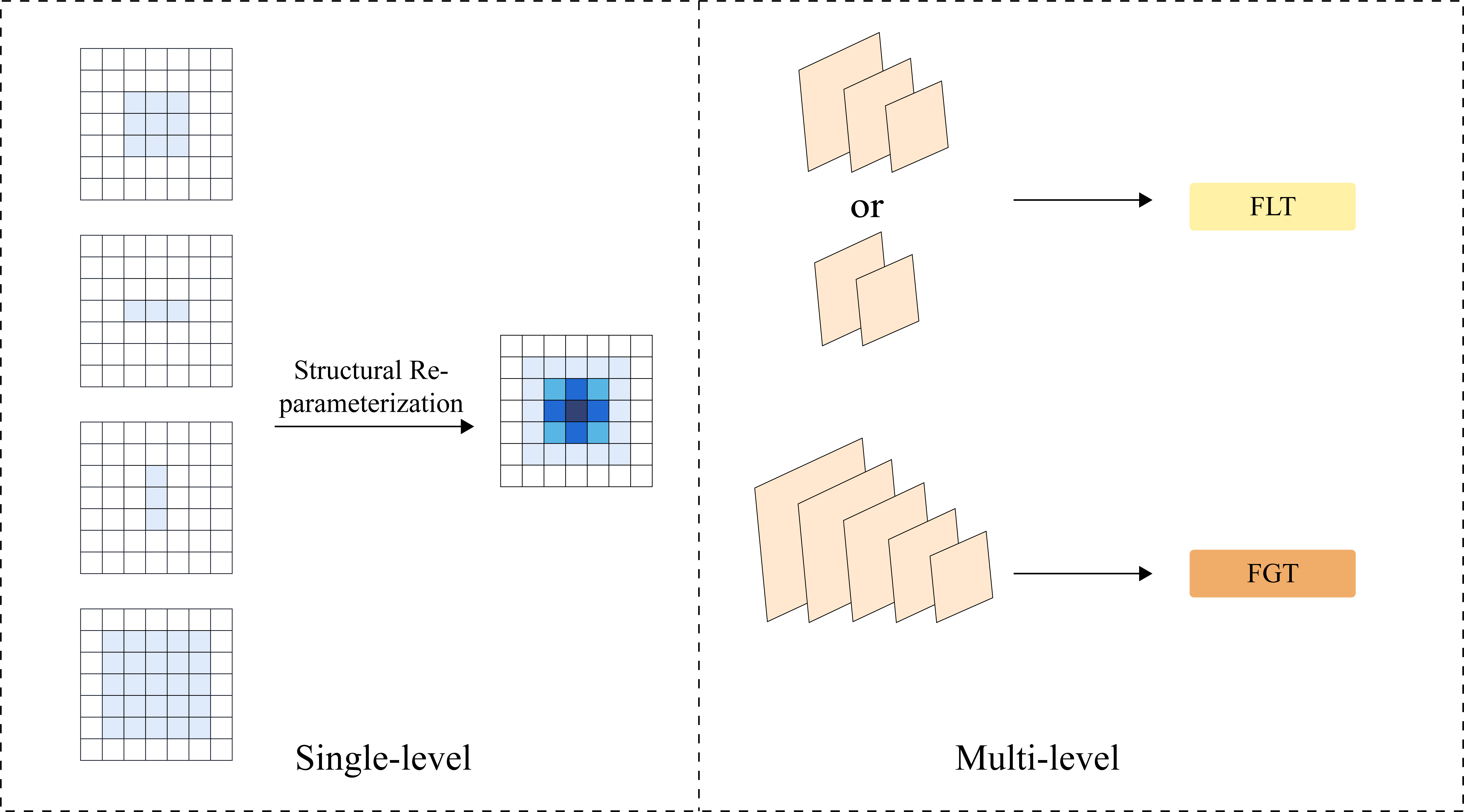}
	\caption{Contrasting two perspectives on multi-scale features. The left figure illustrates the process of multi-scale feature extraction on a single level, while the right figure demonstrates the representation of multi-scale features across multiple levels.}
	\label{Figure 4}
\end{figure}
For the first method, we design a multi-branch module. The module comprises four convolution kernels of varying sizes, each with different receptive fields and sensitivities to target objects of different dimensions. As depicted on the left side of Figure \ref{Figure 4}, this multi-branch structure can be integrated into a single convolution via structural re-parameterization, where darker colors represent greater weights. For the second approach, we introduce a novel topology that utilizes local window self-attention and global self-attention to capture local detail textures and global semantics, respectively. As illustrated on the right side of Figure \ref{Figure 4}, feature pyramids are constructed by selecting feature maps from different stages of the encoder. LFT and GFT modules capture local detail textures and global semantics at different pyramid levels, respectively. These local detail textures and global semantics are aggregated layer by layer through a post-fusion strategy, which merges low-resolution, semantically strong global features with high-resolution, semantically weak local detail textures. By incorporating this topology, LM-Net effectively detects potential lesions in medical images.

\section{Experiments}
\subsection{Datasets}
In this paper, we conduct experiments on three medical image datasets to ensure a comprehensive evaluation of our approach. These datasets include the Kvasir-SEG dataset \cite{Kvasir-seg}, the LGG segmentation dataset \cite{LGG1,LGG2}, and the breast ultrasound image dataset \cite{BUIS}. The Kvasir-SEG dataset comprises 1,000 polyp images, while the LGG segmentation dataset contains 3,929 low-grade glioma images from 110 patients. Lastly, the breast ultrasound image dataset consists of 780 images, which are categorized into normal, benign, and malignant images.
\subsection{Implementation Details}
To ensure fair comparison, we preprocess three medical image datasets using the same pipeline.
Specifcally, each dataset is split into training, validation, and test subsets with a ratio of 0.8:0.1:0.1. 
The input images are resized to $ 256\times256 $ pixels and augmented using the following operations: (1) random horizontal and vertical translations from [-0.9, 1.1] with a probability of 0.5; (2) random scale transformations from [-0.9, 1.1] with a probability of 0.5; (3) random horizontal and vertical flips with a probability of 0.5; (4) random rotations from [$-30^{\circ} $, $30^{\circ} $] with a probability of 0.5; (5) random gaussian blurring with a probability of 0.2; (6)  colorjitter by randomly altering the brightness (factor of 0.2), contrast (0.2), saturation (0.2), and hue (0.2), with a probability of 0.2. These data augmentation strategies are used to increase the diversity of the data. 

For model optimization, we use AdamW \cite{AdamW} with an initial learning rate of 0.001, weight decay of 1e-4, and a cosine annealing schedule for learning rate decay. The batch size is set to 32. The weighted cross-entropy loss function is utilized, where the weighting strategy follows \cite{Balanced_Loss}. Each model is trained for 200 epochs without early stopping. During training, we save the model weights after each epoch if performance (mDice) on the validation set improves, thus avoiding underfitting or overfitting. In our experiments, all models utilized the officially released code.

\subsection{Experiment results}
In this paper, we present LM-Net, a novel medical image segmentation model that outperforms several state-of-the-art methods on three different datasets. LM-Net is compared with various Unet-based methods, including Unet++, Attention Unet, ResUnet, ResUnet++, TransUnet, Swin-Unet and UCTransNet. For fair comparison, we also evaluate some mainstream semantic segmentation models, such as Deeplabv3+ and FCN. To comprehensively evaluate our model, we choose a variety of metrics including accuracy, precision, recall, mean Dice coefficient (mDice) and mean intersection over union(mIoU). In particular, to better assess segmentation boundary quality, we provide Hausdorff distance (Hd) and Relative Area Difference (RAD). Note that Relative Volume Difference (RVD) is commonly used to measure volume differences in 3D, but since our experiments are on 2D images, we simply replace volume with area in the RVD formula to obtain a simplified RAD metric.
Unless explicitly indicated, all experimental results presented in this work reflect the performance on the test sets.
\begin{table}[!ht]
	
	\centering  
	\resizebox{\linewidth}{!}{ 
	\begin{tabular}{cccccccccc}
		\toprule [1pt]
		Model &Accuracy/\%& Precision/\% & Recall/\% & mDice/\% & mIoU/\%  & Hd &RAD/\% & Flops/G & Params/M \\
		\midrule 
		Unet& 95.25±0.08 & 84.36±0.31& 84.92±0.14 & 90.91±0.13& 83.95±0.22& 7.5±0.27&0.95±2.42& 54.74 & 31.03 \\
		Unet++ &96.29±0.08& 88.29±0.78& 87.54±0.4& 92.86±0.12 & 87.07±0.2 & 5.42±0.46&-1.47±0.66& 34.91 & 9.16 \\
		AttUnet&95.82±0.2& 84.6±0.83 & 89.1±0.39 & 92.15±0.35& 85.9±0.56& 6.52±0.44& 6.46±0.72& 66.63 & 34.87 \\
		ResUnet&93.78±0.14& 78.26±1.61& 82.53±1.59& 88.32±0.08& 80±0.14& 8.92±0.4&6.43±1.52& 80.98 & 13.04 \\
		ResUnet++&94.28±0.14& 81.77±0.88& 80.9±0.56&88.98±0.22& 81±0.34& 9.38±0.47&\textbf{-0.61±0.75}& 70.99 & 14.48 \\
		TransUnet &91.06±0.18& 69.11±1.21& 75.87±1.25& 83.5±0.1& 73.26±0.17& 15.36±0.25&11.23±1.4& 38.03 & 87.80 \\
		R50-TransUnet& 96.24±0.12& 88.01±0.64& 87.5±0.16& 92.76±0.2& 86.91±0.32& 5.81±0.07&-4.92±5.62&32.23 & 93.23 \\
		UCTransNet&95.97±0.16& 85.91±1.58 & 88.33±1& 92.35±0.21 & 86.23±0.35& 5.92±0.37& 1.71±2.85&43.06 & 66.24 \\
		Swin-Unet&86.42±0.22& 54.8±0.68 & 67.66±0.74 & 76.17±0.15 & 64.13±0.21 & 24.8±0.38 &29.39±2.97 &11.36 & 41.34 \\
		Deeplabv3+&94.92±0.14 & 84.43±1.98 & 82.22±1.76& 90.15±0.21& 82.78±0.32&7.65±0.82& -2.09±0.3&14.55 & 7.03 \\
		FCN &96.36±0.14 & 86.62±1 & 90.35±0.48 & 93.14±0.24 & 87.53±0.4 &4.49±0.82 &5.86±0.04 &34.71 & 32.94 \\
		LM-Net&\textbf{96.91±0.07}&	\textbf{89.64±0.58}&\textbf{90.38±0.23}&\textbf{94.09±0.11}&\textbf{89.12±0.19}&	\textbf{2.76±0.12}&1.53±1.4&\textbf{4.66} &\textbf{5.40} \\
		\bottomrule [1pt] 
	\end{tabular}
}
	\caption{The table shows the comprehensive evaluation results of some models on the Kvasir-SEG test dataset.}
	\label{Table1}
\end{table}
\begin{table}[!ht]
	\centering  
	\resizebox{\linewidth}{!}{ 
	\begin{tabular}{cccccccc}
		\toprule [1pt]
		Model &Accuracy/\%& Precision/\% & Recall/\% & mDice/\% & mIoU/\% & Hd &RAD/\% \\
		\midrule 
		Unet &99.79±0.01& 89.33±0.65& 91.42±0.59& 95.13±0.06 & 91.1±0.11& 0.48±0.15 & 1.7±0.58\\
		Unet++ &99.8±0.02&90.83±0.22& 91.07±0.11& 95.43±0.02 & 91.6±0.05 &0.36±0.12& \textbf{-0.23±0.14} \\
		AttUnet &99.79±0.01& 90.07±0.18& 91.06±0.21 &95.23±0.01 &91.27±0.03&0.58±0.02&-3.66±4.84\\
		ResUnet&99.75±0.02& 89±0.48& 87.76±0.56& 94.12±0.06 & 89.46±0.09&0.46±0.03&-1.4±0.07 \\
		ResUnet++ &99.78±0.01& 89.87±1.42& 89.74±1.15 & 94.84±0.07& 90.63±0.11&\textbf{0.33±0.03}& 1.4±0.66\\
		TransUnet &99.79±0.01&90.77±0.29& 89.93±0.32 & 95.12±0.03 & 91.09±0.05&0.37±0.01&-0.97±0.23\\
		R50-TransUnet &99.8±0.01& \textbf{90.84±0.25}& 90.75±0.17& 95.35±0.02& 91.47±0.04&0.39±0.01&-0.44±0.1\\
		UCTransNet &99.8±0.02& 90.34±0.25& 91.51±0.14& 95.41±0.03& 91.58±0.05&0.39±0.03&1.71±0.01\\
		Swin-Unet &99.68±0.03& 87.08±0.22& 82.71±0.4& 92.34±0.07& 86.67±0.11 &0.74±0.04& -5.15±0.32\\
		Deeplabv3+ &99.8±0.01& 90.43±0.68& \textbf{91.63±0.76} &95.46±0.03 & 91.67±0.05 & 0.41±0.08&1.68±0.01\\
		FCN &99.77±0.01& 88.69±0.03& 90.69±0.06& 94.78±0.02&90.53±0.02& 0.37±0.03&2.38±0.12\\
		LM-Net&\textbf{99.81±0.01} & 90.55±0.76 & 91.58±0.65 & \textbf{95.48±0.07} & \textbf{91.7±0.13} &0.36±0.01&2.67±0.05\\
		\bottomrule [1pt] 
	\end{tabular}
}
	\caption{The table shows the experimental result on the LGG test dataset.}
	\label{Table2}
\end{table}
\begin{table}[!ht]
	\centering 
	\resizebox{\linewidth}{!}{  
	\begin{tabular}{cccccccc}
		\toprule [1pt]
		Model & Accuracy/\% & Precision/\% & Recall/\% & mDice/\% & mIoU/\% &Hd&RAD/\%\\
		\midrule 
		Unet  & 96.45±0.1& 85.28±1.59& 76.21±1.8& 89.26±0.42& 81.74±0.6&7.32±0.83&-13.04±1\\
		Unet++ &96.54±0.1&\textbf{ 87.39±2.13} & 74.76±1.12& 89.34±0.14& 81.87±0.22 &7.52±2.74 &-13.29±1.77\\
		AttUnet & 96.64±0.09& 86.3±1.48& 77.24±0.64 & 89.83±0.18& 82.58±0.26 &7.26±0.81&-9.51±1.02\\
		ResUnet & 94.78±0.05& 78.08±0.73& 63.56±0.5& 83.59±0.11 & 74.17±0.13& 15.61±0.55 &-17.6±2.11\\
		ResUnet++& 96.44±0.04& 86.64±1.03& 74.4±0.63 & 89.05±0.06& 81.45±0.08&8.56±0.27&-14.04±0.57\\
		TransUnet & 94.72±0.11 & 78.76±1.82 & 61.67±1.49& 83.14±0.41 & 73.62±0.5&15.12±0.42&-24.18±3.04\\
		R50-TransUnet& 96.49±0.11& 84.88±1.44 & 77.32±2.17 & 89.47±0.36 & 82.05±0.52&6.09±0.38&-11±1.12\\
		UCTransNet& 96.7±0.09& 84.55±1.76& 80.38±1.31& 90.29±0.11& 83.25±0.17&6.08±0.74&\textbf{-3.56±0.68}\\
		Swin-Unet&  91.49±0.31& 57.11±2.74& 46.96±3.58& 73.39±0.38& 62.86±0.25&27.68±1.21 &-26.64±3.46\\
		Deeplabv3+ &96.43±0.16& 84.99±1.45 & 76.33±0.24& 89.23±0.41& 81.7±0.61& 7.44±0.98&-11.61±1.36\\
		FCN & 96.24±0.14& 81.23±2.02& 79.13±1.13 & 89.04±0.23 & 81.41±0.34&8.42±0.94& -4.28±0.03\\
		LM-Net & \textbf{96.91±0.08} &85.03±1.49&\textbf{82.29±1.11} & \textbf{90.96±0.24} & \textbf{84.26±0.37}&\textbf{5.59±1.14}&-6.36±4.31\\
		\bottomrule [1pt] 
	\end{tabular}
}
	\caption{The table shows the experimental result on the breast ultrasound test dataset.}
	\label{Table3}
\end{table}
\begin{table}[!ht]
	\centering  
	\resizebox{\linewidth}{!}{ 
	\begin{tabular}{c|c|cccc|cccc}
		\hline
		& Normal & \multicolumn{4}{c|}{Benign} &\multicolumn{4}{c}{Malignant}  \\
		\hline
		Model & Accuracy/\% & mDice/\% & mIoU/\% & Hd&RAD/\% & mDice/\% & mIoU/\% & Hd &RAD/\%\\
		\hline
		Unet &99.26±0.02&89.27±0.59&81.83±0.85&	9.31±1.22&	-11.52±1.62&87.31±0.55&	78.54±0.79&	7.2±0.32&-17.26±0.43\\
		Unet++ &99.93±0.04&90.15±0.34&	83.09±0.5&	9.14±0.04&	-10.27±0.38& 87.04±0.45&	78.13±0.62&	10.48±1.63&	-16.69±3.09\\
		AttUnet &\textbf{99.94±0.02} &90.12±0.01&	83.03±0.01&	10.45±1.37&	-6.73±0.33&88.27±0.22&	79.9±0.31&	9.29±0.44&	-12.65±1.79\\
		ResUnet &99.67±0.1&85.32±0.14&	76.41±0.19&	18.77±1.78&	-9.11±2.27&80.19±0.29&	69.19±0.34&	13.09±0.07&	-27.66±1.62\\
		ResUnet++ & 99.48±0.04& 89.02±0.42&	81.46±0.59&	10.69±1.03&	-13.02±1.61&87.97±0.19&	79.48±0.28&	6.8±0.05&	-16.96±0.64\\
		TransUnet & 99.49±0.22& 83.49±0.32&	74.15±0.41&	15.07±0.79&	-20.1±2.98&81±0.18&	70.21±0.19&	10.26±0.19&	-30.26±2.31\\
		R50-TransUnet & 99.53±0.07 &87.52±0.48&	79.36±0.65&	9.74±0.53&	-14.18±1.29&89.91±0.83&	82.36±1.28&	5.01±0.46&	\textbf{-1.19±0.66}\\
		UCTransNet & 99.19±0.31&90.04±0.16&	82.92±0.23&	\textbf{7.29±0.71}&	-8.61±0.34&90.17±0.17&	82.73±0.26&	6.5±0.43&-9.35±3.39	\\
		Swin-Unet & 99.1±0.3&72.69±0.03&	62.23±0.08&	32.18±1.25&	\textbf{-4.1±3.49}&72.47±0.43&	60.98±0.39&	13.95±0.16&	-53.39±2.41\\
		Deeplabv3+ &99.08±0.09& 88.39±0.26&	80.56±0.35&	11.35±0.99&	-15.04±1.2&89.33±0.23&	81.47±0.36&	6.96±0.28&	-11.29±1.82\\
		FCN & 99.4±0.23&  87.88±0.5&79.84±0.68&	10.67±0.01&	-8.72±2.82&88.66±0.08&	80.44±0.11&	7.33±0.12&	-1.78±3.7\\
		LM-Net & 98.72±0.2&\textbf{90.46±0.02}&	\textbf{83.54±0.03}&	7.18±0.27&	-9.68±1.76 &\textbf{90.86±0.61}&	\textbf{83.8±0.96}&	\textbf{4.75±0.12}&	-7.35±6.18\\
		\hline
	\end{tabular}
}
	\caption{The table shows the experimental result on the breast ultrasound test dataset.}
	\label{Table4}
\end{table}
\begin{table}[!ht]
	\centering  
	\begin{tabular}{ccccc}
		\toprule [1pt]
		Model & mDice/\% & mIoU/\% & Flops/G & Params/M \\
		\midrule 
		FCBFormer \cite{FCBFormer} & 93.85 & 89.03 & 38.77 & 52.94 \\
		SSFormer-L \cite{SSFormer} & 93.57 & 89.05 & 17.28 & 65.95 \\
		ESFPNet-L \cite{ESFPNet} & 92.5 &87.5 & 11.60 & 61.69 \\
		LM-Net					 & \textbf{94.09±0.11}&\textbf{89.12±0.19} & \textbf{4.66} & \textbf{5.40} \\
		\bottomrule [1pt] 
	\end{tabular}
	\caption{ The table shows a comparison with state-of-the-art methods on the Kvasir-SEG test dataset.}
	\label{Table5}
\end{table}
\begin{figure}[!ht]
	\centering
	\includegraphics[width=1\linewidth]{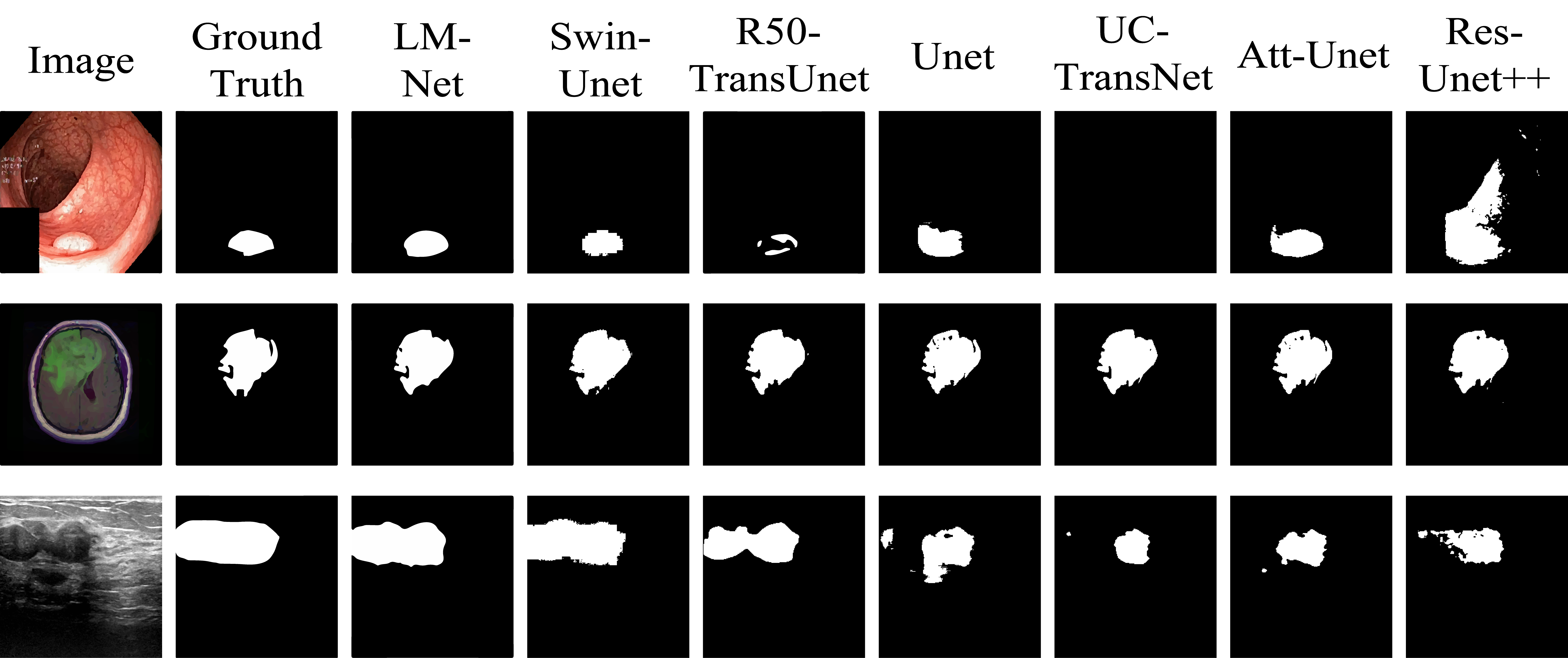}
	\caption{Comparing the qualitative results of various models, we discovered that pure CNN-based methods tend to over-segment, whereas Transformer-based methods are more prone to under-segmentation.}
	\label{Figure 5}
\end{figure}
\begin{figure}[!ht]
	\centering
	\includegraphics[width=1\linewidth]{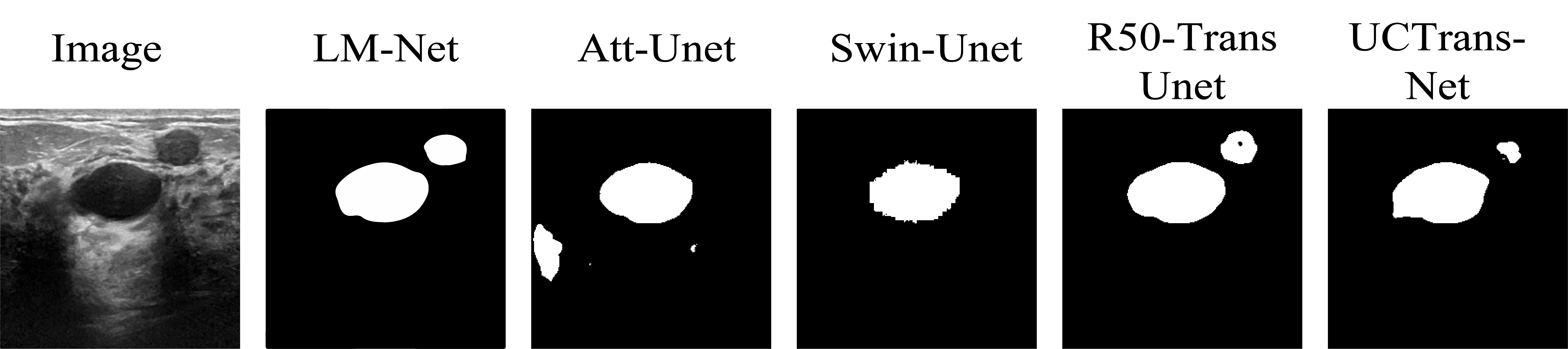}
	\caption{Comparison results of LM-Net and other best-performing models. LM-Net can effectively detect potential targets and produce smoother segmentation boundaries.}
	\label{Figure 6}
\end{figure}

On the Kvasir-SEG dataset, LM-Net achieves the highest accuracy, precision, mDice, mIoU and lowest Hausdorff distance (Hd), significantly improving over prior arts with only 4.66G FLOPs and 5.4M parameters as shown in Table \ref{Table1}.
Additionally, LM-Net demonstrates effective feature extraction on the LGG segmentation dataset, attaining state-of-the-art mDice of 95.48\% and mIoU of 91.7\% as presented in Table \ref{Table2}. To further evaluate the performance of LM-Net, we test it on the breast ultrasound dataset, which contains three categories: normal, benign, and malignant. Benign samples usually have small tumors, while malignant samples usually have blurry boundaries, each with distinct characteristics that can better evaluate the representation learning ability of the model. As shown in Table \ref{Table3}, LM-Net again achieves the best results, demonstrating its superior representation learning ability. We further analyze per-class results on the ultrasound dataset in Table \ref{Table4}. Unlike R50-TransUnet which is more sensitive to malignant tumors but performs poorly on benign cases, and UCTransNet which focuses more on benign tumors, LM-Net can make accurate judgments for all classes.
Finally, we compare LM-Net with state-of-the-art methods on the Kvasir-SEG dataset. The results, reported in Table \ref{Table5}, show that LM-Net outperforms existing models, such as ESFPNet-L \cite{ESFPNet}, and achieves competitive results with FCBFormer \cite{FCBFormer} and SSFormer.\cite {SSFormer}.

\begin{figure}
	\centering
	\includegraphics[width=1.0\linewidth]{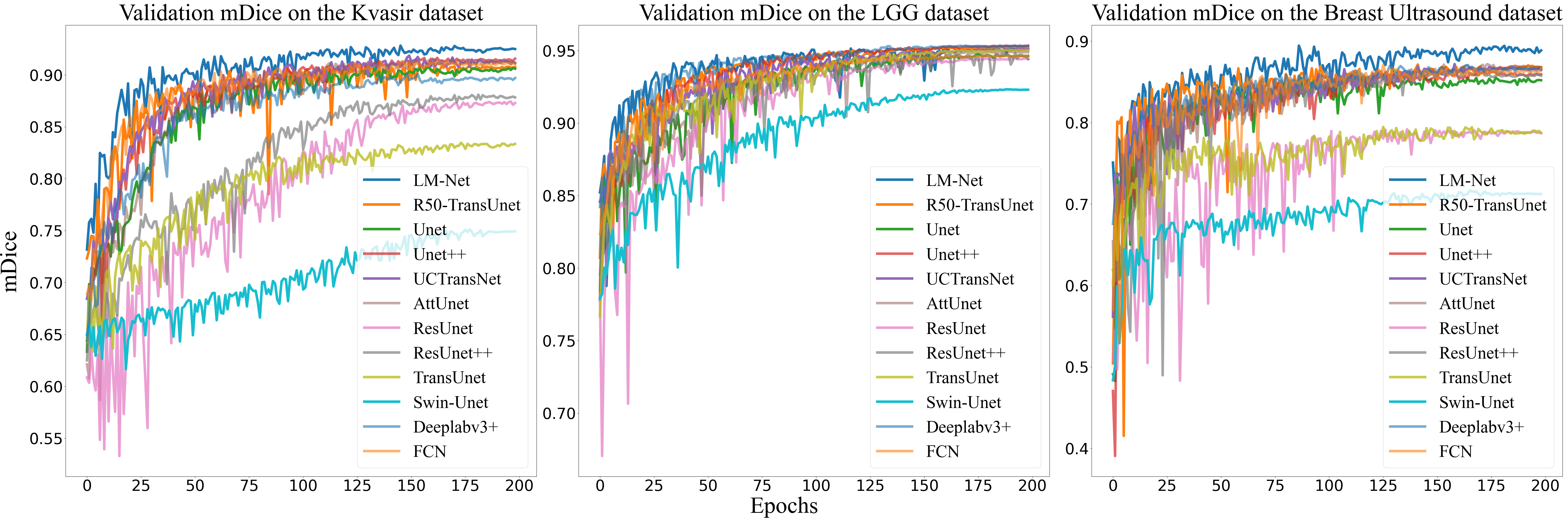}
	\caption{Visualize the trend of mDice on the validation set across epochs for each model.}
	\label{Figure 7}
\end{figure}
The comparative analysis and qualitative results for various segmentation models are presented in Figure \ref{Figure 5}. Our findings indicate that pure CNN-based methods tend to over-segment or under-segment. For example, Unet exhibits over-segmentation on the polyp dataset, characterized by lower precision than recall. On the breast ultrasound dataset, it appears under-segmented with precision surpassing recall. In contrast,  Transformer-based methods tend to be under-segmented. For example, R50-TransUnet exhibits under-segmentation on both datasets with higher precision than recall. What we need to clarify is that this observation is only relative, since different models demonstrate varying generalization capabilities and parameter complexity.
 As can be seen from the visualized image, LM-Net produces smoother edges compared to other methods, quantified by its lower Hausdorff distance of 2.76 on Kvasir and 5.59 on the ultrasound data. Further comparison between LM-Net and other best-performing models (as indicated in Table \ref{Table4}) in Figure \ref{Figure 6} indicates LM-Net's superior ability in capturing potential semantic information. These results demonstrate LM-Net can perform more fine-grained segmentation while preserving shape information.  
 
 Importantly, the proposed GFT and LFT modules do not require extensive pre-training. As visualized in the validation mDice curves over epochs in Figure \ref{Figure 7}, our LM-Net model skews towards the upper-left corner, demonstrating faster convergence compared to other approaches. This efficiency is attributed to LM-Net's effective integration of local and global representations, which is further supported by our ablation experiments. The faster convergence indicates LM-Net's ability to optimize effectively without reliance on heavy pre-training, owing to the complementary design that combines multi-scale and multi-perspective feature learning.

\subsection{Ablation Studies}
\textbf{Multi-Branch Module}. 
As reported in Table \ref{Table6}, the impact of the multi-branch module on the performance of LM-Net is evaluated. Specifically, we compare the effect of expanding the convolutional kernel size and constructing a parallel branch consisting $ 5\times 5 $ and $ 3\times 3 $ convolutions. The results show that the parallel branch structure may be a better choice than simply expanding the convolution kernel. For instance, simply replacing a $3\times3$ convolution with a larger $5\times5$ convolution leads to minor performance gains, with mDice increased by approximately 0.69\% and mIoU increased by 1.15\%. However, by constructing parallel branches with both $5\times5$ and $3\times3$ convolutions, more significant improvements can be achieved, with the mDice improved by 0.78\% and mIoU improved by 1.29\%. These results demonstrate that utilizing larger and smaller convolutions in parallel, rather than naively enlarging convolution kernels, can more effectively improve model generalization capability. 
 Inspired by asymmetric convolution, we also explore the impact of using asymmetric convolutions. 
Our results show that a multi-branch structure combining large convolutions in parallel with small convolutions and asymmetric convolutions can yield significant improvements.
This is attributed to the fact that convolutions with different receptive fields can capture features at different scales. Additionally, we explore the influence of additional branches and larger convolutional kernels on the model's performance. Interestingly, it is observed that the model's performance declines, potentially due to an excess of redundant information.
\begin{table}[!ht]
	\centering  
	\resizebox{\linewidth}{!}{ 
	\begin{tabular}{cccccccccc}
		\toprule [1pt]
		3×3 & 5×5 & 1×3 & 3×1 & Dilation5×5	& 7×7 & mDice/\% & mIoU/\%&Hd&RAD/\% \\
		\midrule 
		\checkmark &  &  &  &  &  &92.86±0.34&	87.06±0.57&	4.43±0.66&	5.27±3.62\\
		& \checkmark &  &  &  &  & 93.55±0.34&	88.21±0.58&	4.12±0.32&	2.16±2.31\\
		&  & \checkmark & \checkmark &  &  & 92.56±0.2&	86.58±0.32&	5.96±0.13&	3.66±2.23\\
		\checkmark & \checkmark &  &  &  &  & 93.64±0.09&	88.35±0.14&	4.19±0.55&	4.56±1.99\\
		\checkmark &  & \checkmark & \checkmark &  &  & 92.87±0.17&	87.09±0.29&	4.89±0.38&	\textbf{1.1±4.07}\\
		& \checkmark & \checkmark & \checkmark &  &  & 93.6±0.11&	88.29±0.17&	4.32±0.12&	1.23±2.76\\
		\checkmark & \checkmark & \checkmark & \checkmark &  &  & \textbf{94.09±0.11}&	\textbf{89.12±0.19}&	\textbf{2.76±0.12}&	1.53±1.4\\
		\checkmark & \checkmark & \checkmark & \checkmark & \checkmark &  & 93.47±0.34&	88.08±0.58&	3.45±0.28&	3.04±0.94\\
		\checkmark & \checkmark & \checkmark & \checkmark & \checkmark & \checkmark & 93.85±0.17&	88.71±0.28&	3.46±0.41&	2.8±1.68\\
		\bottomrule [1pt] 
	\end{tabular}
}
	\caption{The effects of convolution combinations of different scales are investigated.}
	\label{Table6}
\end{table}
\begin{table}[!ht]
	\centering  
	\begin{tabular}{ccc}
		\toprule [1pt]
		structural re-parameterization & Flops/G & Params/M \\
		\midrule 
		& 5.15 & 5.45 \\
		\checkmark & \textbf{4.66} & \textbf{5.40} \\
		\bottomrule [1pt] 
	\end{tabular}
	\caption{The influence of structural re-parameterization on the calculation amount and parameter amount of the model.}
	\label{Table7}
\end{table}
Table \ref{Table7} presents the benefits of structural re-parameterization. With this technique, the computational cost is reduced by approximately 0.5G FLOPs while also decreasing model parameters. Our results demonstrate that structural re-parameterization can effectively enhance model efficiency by reducing computational and parameter complexity without compromising performance.
\\
\textbf{GFT}. In this section, a comprehensive comparison in Table \ref{Table8} verifies the rationale behind our GFT design. GFT reduces the Hausdorff distance by about 1 point compared to ASPP, with only 1.56M additional parameters.
We then discuss the design of the GFT module. TransUnet utilizes the deep abstraction feature of the encoder as input of the Transformer, which is a single-scale representation. In contrast, the feature pyramid was empirically chosen as the input of the Transformer in our approach, which can exploit the multi-scale features at different levels to perceive the global contextual information.
 An experiment was conducted to compare multi-scale representation with single-scale representation: The multi-scale representation stacks features of different scales from the encoder to build a global feature pyramid as input of the Transformer, while the single-scale representation chooses the last feature from the encoder as input of the Transformer. To mitigate the impact of inconsistent channel dimensions, a $ 1\times1 $ convolution layer was appended to expand channel dimensions of the last feature in the single-scale representation to match the pyramid. Experiments comparing the multi-scale pyramid versus single-scale encoder features show the former as a superior design choice for Transformer input. 
\begin{table}[!ht]
	\centering 
	\resizebox{\linewidth}{!}{  
\begin{tabular}{cccccccccc}
	\toprule [1pt]
	&Accuracy/\%& precision/\% & recall/\% & mDice/\% & mIoU/\%&Hd&RAD/\% & Flops/G & Params/M \\
	\midrule 
	Baseline & 96.33±0.14&	86.44±0.2&	90.4±0.71&	93.09±0.27&	87.44±0.45&	4.48±0.18&	3.91±1.59
	 &4.04 & 1.13 \\
	ASPP &96.65±0.26&	88.55±2.32&	89.92±1.02&	93.62±0.41&	88.32±0.7&	3.79±0.19&	\textbf{0.01±2.18}
	& 4.73 & 3.89 \\
	Single-scale  &96.62±0.1&	87.78±0.87&	\textbf{90.66±0.38}&	93.6±0.16&	88.28±0.28&	4.69±0.34&	-0.29±4.37
	& 5.18 & 5.57 \\
	Multi-scale  & \textbf{96.91±0.07}&	\textbf{89.64±0.58}&	90.38±0.23&	\textbf{94.09±0.11}&	\textbf{89.12±0.19}&	\textbf{2.76±0.12}&	1.53±1.4& 5.15 & 5.45 \\
	\bottomrule [1pt] 
\end{tabular}
}
	\caption{Ablation study on the GFT module. The Baseline means no other operation is performed, and the output of the encoder is directly used as the input of the decoder. Single-scale  indicates taking the last feature of the encoder as input of the Transformer. Multi-scale denotes taking stacked features from different scales to build a global feature pyramid as input of the Transformer.}
	\label{Table8}
\end{table}
\\
\textbf{LFT}. This section investigates the impact of LFT as a multi-scale skip connection (Multi-skip). The original Unet uses a single-scale skip connection (Single-skip) that fails to effectively utilize multi-scale features. In contrast, our proposed multi-skip connects the encoder and decoder via the feature pyramid to incorporate multi-scale information. As shown in Table \ref{Table9}, multi-skip alone outperforms single-skip in most metrics, validating the importance of multi-scale features in skip connections. Further, combining multi-skip and LFT modules achieves additional performance gains, demonstrating their complementary benefits. Overall, the results highlight the significance of integrating multi-scale representations in skip connections to enhance generalization for semantic segmentation models.
\begin{table}[!ht]
	\centering  
	\resizebox{\linewidth}{!}{ 
\begin{tabular}{ccccccccc}
	\toprule [1pt]
	Single-skip & Mulit-skip & LFT & mDice/\% & mIoU/\%&Hd&RAD/\% & Flops/G & Params/M \\
	\midrule 
	\checkmark &  &  & 92.84±0.53&	87.03±0.87&	5.36±0.42&	7.24±0.41
	& 3.61 & 4.98 \\
	\checkmark &  & \checkmark & 93.25±0.13&	87.7±0.22&	4.12±0.16&	3.95±1.11
	& 4.77 & 5.35 \\
	& \checkmark &  & 93.13±0.31&	87.5±0.52&	4.05±0.33&	4.42±1.93
	& 3.98 & 5.07 \\
	& \checkmark & \checkmark & \textbf{94.09±0.11}&	\textbf{89.12±0.19}&	\textbf{2.76±0.12}&\textbf{	1.53±1.4}
		& 5.15 & 5.45 \\
	\bottomrule [1pt] 
\end{tabular}
}
	\caption{Ablation study on LFT module. The Single-skip represents a simple skip-connection scheme in Unet. Multi-skip means selecting outputs from adjacent stages of the encoder to construct a pyramid as a skip-connection scheme.}
	\label{Table9}
\end{table}
\begin{table}[!ht]
	\centering 
	\resizebox{\linewidth}{!}{  
\begin{tabular}{cccccccc}
	\toprule [1pt]
	Baseline & Multi-Branch & LFT & GFT & mDice/\% & mIoU/\%&Hd&RAD/\% \\
	\midrule 
	\checkmark &  &  &  & 90.91±0.13 &	83.95±0.22 &	7.5±0.27 &	0.95±2.42\\
	\checkmark & \checkmark &  &  & 92.24±0.11&	86.08±0.2&	4.17±0.11&	\textbf{0.22±3.76}\\
	\checkmark & \checkmark & \checkmark &  & 93.09±0.27&	87.44±0.45&	4.48±0.18&	3.91±1.59\\
	\checkmark & \checkmark &  & \checkmark & 92.84±0.53&	87.03±0.87&	5.36±0.42&	7.24±0.41\\
	\checkmark & \checkmark & \checkmark & \checkmark &\textbf{94.09±0.11}&\textbf{89.12±0.19}&\textbf{2.76±0.12}&	1.53±1.4
	 \\
	\bottomrule [1pt] 
\end{tabular}
}
	\caption{Comparing the impact of each module. The Baseline represents Unet. The robust result indicates that it is essential to combine the local detail texture and the global semantic information for dense prediction tasks.}
	\label{Table10}
\end{table}
\\
\textbf{Overall}. Table \ref{Table10} presents the impact of each module on model performance using Unet as the baseline. Adding the multi-branch module improves mIoU by about 2.13\%, again showing its efficacy in extracting multi-scale features to boost generalization. Further incorporating LFT or GFT enhances representational learning. Combining LFT and GFT achieves the best robustness, highlighting the importance of integrating local textures and global semantics for dense prediction tasks.
\section{Discussion and Conclusion}
In this paper, we propose LM-Net, a lightweight and effective U-shaped architecture for medical image segmentation, integrating strengths of CNNs and Vision Transformers. The design incorporates three key modules: the multi-branch module, Local Feature Transformer (LFT), and Global Feature Transformer (GFT). It explores multi-scale representations from two perspectives: the multi-branch module extracts multi-scale features at a same level, whereas LFT and GFT capture multi-scale information at different levels. Concurrently, LFT employs local window self-attention to capture fine-grained textures, while GFT utilizes global self-attention to model global contexts. The decoder then aggregates these local and global multi-scale representations complementarily, enabling the model to learn meaningful anatomical semantics for accurate segmentation. By combining CNNs and Transformers to model textures, context, and multi-scale information, LM-Net demonstrates robust feature extraction capabilities for precise region of interest delineation.

Compared to other medical image segmentation models, our proposed LM-Net  achieves state-of-the-art results on three publicly available datasets. This success can be attributed to two main advantages: (1) LM-Net effectively combines local detail textures and global semantic information to improve its representation ability. As shown in Table \ref{Table10}, the improvement in the model's representational capacity is primarily due to the integration of LFT and GFT. This complementary approach enables the model to perceive potential semantic information and overcome the problem of ambiguous segmentation boundaries. (2) Multi-scale features are explored from different perspectives. As evidenced in Tables \ref{Table6}, \ref{Table8}, and \ref{Table9}, our proposed multi-perspective multi-scale representation significantly improves model generalization, even surpassing ASPP.

In designing the multi-branch module, we found excessive parallel branches degrade model generalization. This is because too many branches generate redundant information, which impedes the model's training and results in a suboptimal solution. This hypothesis is supported by previous research \cite{Repvgg,Acnet,DBB} and further confirmed in Table \ref{Table6}. Therefore, we carefully selected the branch number to balance complexity and performance. Additionally, LM-Net was randomly initialized, which may be an inferior scheme. The parameters of LM-Net primarily focus on GFT, so we are considering using some variants of Transformer to further reduce the parameters and calculation, such as CCT \cite{CCT}.

In conclusion, our proposed LM-Net effectively tackles key limitations in medical image segmentation including over-segmentation, under-segmentation and blurred boundaries. Through its multi-scale design, the model strikes an optimal balance between accuracy and efficiency. Comprehensive experiments validate that employing the feature pyramid for global context modeling and local detail extraction in the Transformer, followed by effective integration, helps alleviate segmentation ambiguity. The consistent state-of-the-art results across multiple datasets highlight the significance of multi-perspective multi-scale feature learning for advanced medical image analysis.

\section{Acknowledgement}
This work was supported in part by the National Natural Science Foundation of China under Grants 62071382, 82030047 and 61561008, as well as the Guangxi Science and Technology Major Special Project (2023AA20012). 
\bibliographystyle{unsrt}
\bibliography{document.bib}

\end{document}